\documentclass[letterpaper]{article} 
\usepackage{aaai23}  
\usepackage{times}  
\usepackage{helvet}  
\usepackage{courier}  
\usepackage[hyphens]{url}  
\usepackage{graphicx} 
\usepackage{natbib}  
\usepackage{caption} 
\usepackage{algorithm}
\usepackage{algorithmic}

\usepackage{newfloat}
\usepackage{listings}
\DeclareCaptionStyle{ruled}{labelfont=normalfont,labelsep=colon,strut=off} 
\floatstyle{ruled}
\newfloat{listing}{tb}{lst}{}
\floatname{listing}{Listing}
\usepackage{amsmath,amssymb}
\usepackage{array}
\usepackage{booktabs}
\usepackage{multirow}

\title{Breaking Immutable: Information-Coupled \\ Prototype Elaboration for Few-Shot Object Detection}
\author{
    Xiaonan Lu\textsuperscript{\rm 1,\rm 2,\rm 3,\rm 4},
    Wenhui Diao\textsuperscript{\rm 1,\rm 2,\rm 3,\rm 4}\thanks{Corresponding author.},
    Yongqiang Mao\textsuperscript{\rm 1,\rm 2,\rm 3,\rm 4},
    Junxi Li\textsuperscript{\rm 1,\rm 2,\rm 3,\rm 4},
    Peijin Wang\textsuperscript{\rm 1,\rm 2},\\
    Xian Sun\textsuperscript{\rm 1,\rm 2,\rm 3,\rm 4},
    Kun Fu\textsuperscript{\rm 1,\rm 2,\rm 3,\rm 4}
}
\affiliations{
    \textsuperscript{\rm 1}Aerospace Information Research Institute, Chinese Academy of Sciences\\
    \textsuperscript{\rm 2}Key Laboratory of Network Information System Technology (NIST), Aerospace Information Research Institute, \\Chinese Academy of Sciences
    \textsuperscript{\rm 3}University of Chinese Academy of Sciences\\
    \textsuperscript{\rm 4}School of Electronic, Electrical and Communication Engineering, University of Chinese Academy of Sciences\\
    \{luxiaonan19, maoyongqiang19, lijunxi191, wangpeijin17\}@mails.ucas.ac.cn\\
    \{diaowh, sunxian\}@aircas.ac.cn, kunfuiecas@gmail.com
}

\usepackage{bibentry}

\begin{document}

\maketitle

\begin{abstract}
Few-shot object detection, expecting detectors to detect novel classes with a few instances, has made conspicuous progress. However, the prototypes extracted by existing meta-learning based methods still suffer from insufficient representative information and lack awareness of query images, which cannot be adaptively tailored to different query images. Firstly, only the support images are involved for extracting prototypes, resulting in scarce perceptual information of query images. Secondly, all pixels of all support images are treated equally when aggregating features into prototype vectors, thus the salient objects are overwhelmed by the cluttered background. In this paper, we propose an Information-Coupled Prototype Elaboration (ICPE) method to generate specific and representative prototypes for each query image. Concretely, a conditional information coupling module is introduced to couple information from the query branch to the support branch, strengthening the query-perceptual information in support features. Besides, we design a prototype dynamic aggregation module that dynamically adjusts intra-image and inter-image aggregation weights to highlight the salient information useful for detecting query images. Experimental results on both Pascal VOC and MS COCO demonstrate that our method achieves state-of-the-art performance in almost all settings.
\end{abstract}

\begin{figure}[!t]
	\centering
	\includegraphics[width=0.85\columnwidth]{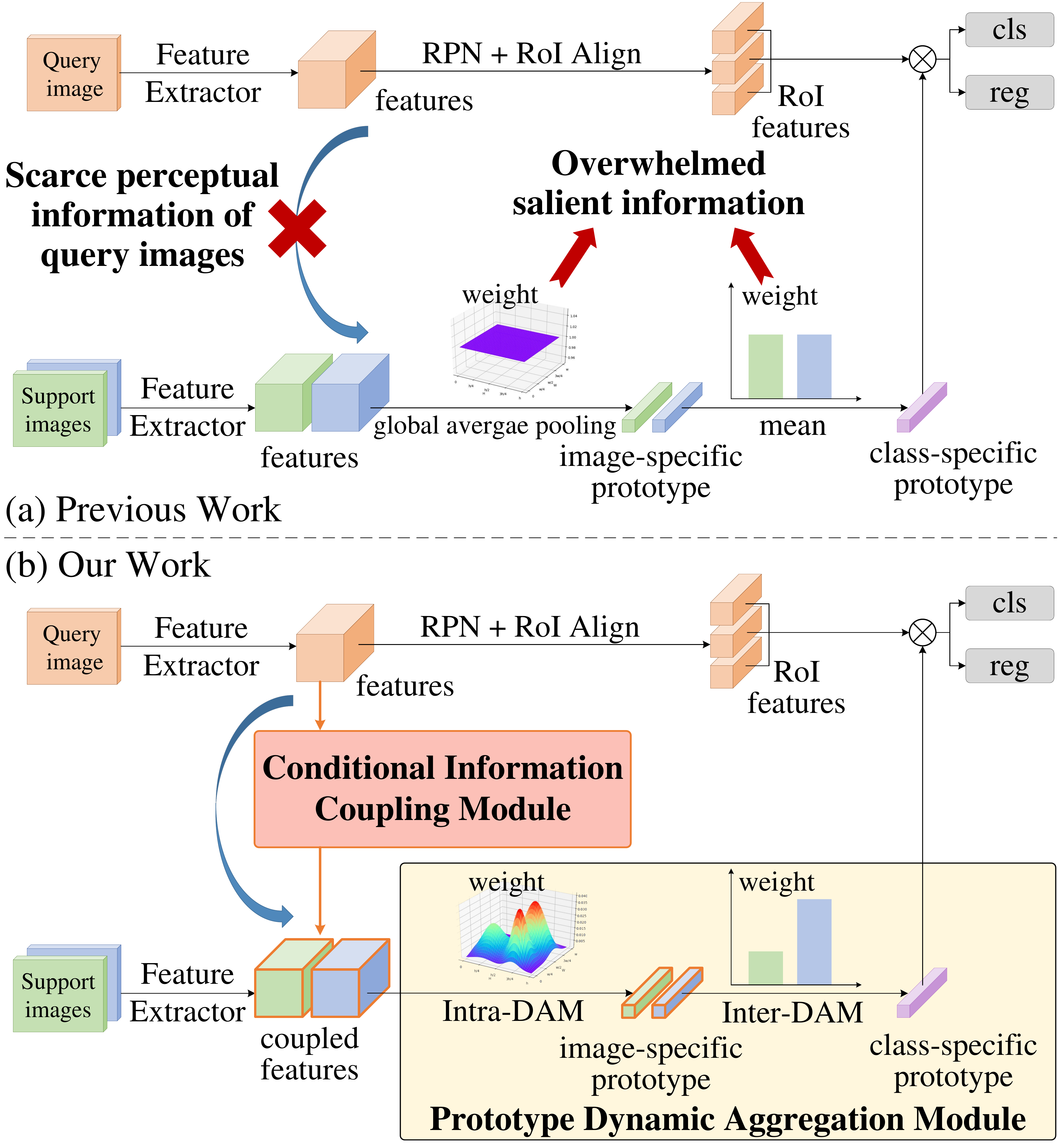}
	\caption{Prototype extraction in meta-learning methods. (a) Prototypes extracted in previous works lack perceptual information of query images. And averagely aggregating features makes the salient information overwhelmed. (b) In our work, a conditional information coupling module strengthens query-perceptual information. And we design a prototype dynamic aggregation module consisting of intra- and inter-image dynamic aggregation mechanisms (Intra-DAM, Inter-DAM) to consolidate salient information of prototypes.}
	\label{Figure 1}
\end{figure}

\section{Introduction}

Object detection has witnessed significant progress by deep learning \cite{ren2015faster,redmon2016you,lin2017focal,carion2020end}. However, most detectors require abundant labeled data for training and impracticably generalize to new tasks with a few images. In contrast, humans can observe novel objects with limited instances. Thus, few-shot object detection (FSOD) is proposed to bridge the gap between deep learning and human learning systems.

FSOD assists in improving the performance on novel classes with a few labeled images by learning knowledge on base classes with abundant data. Some methods \cite{chen2018lstd,wang2020frustratingly} transfer the knowledge from base dataset to novel dataset. Still, they are sensitive to the difference between datasets and may produce negative transfer when the difference is obvious. Other methods \cite{hsieh2019one,yan2019meta,li2021beyond} achieve remarkable performance based on meta-learning. They simulate few-shot scenes and construct episodes consisting of support sets and query sets. The annotated support sets are used to generate prototypes, then the objects in query sets are detected based on the generated prototypes. The quality of prototypes directly affects the performance on query images. However, the prototypes extracted in most meta-learning based methods are immutable, suffering from insufficient query-aware and representative information. They are unable to be adaptively tailored to different query images. As Figure~\ref{Figure 1}(a) shows, it mainly manifests in the following two aspects:

\begin{enumerate}
	\item[1)] \textit{Scarce perceptual information of query images.} Existing methods extract prototypes only from support images, lacking the perception of query images. The prototypes extracted from the same support images are the same, which may be not suitable for predicting all query images. It is inevitable that the detection performance of different query images varies greatly, especially when the extreme perspective span occurs.
	
	\item[2)] \textit{Overwhelmed salient information of support images.} Majority works averagely aggregate support features into prototypes. All pixels of all images contribute equally to prototypes. On the one hand, using averaging operation within an image (intra-image) causes that the cluttered background weakens the interested objects. On the other hand, the similarities between support images and query images are different. Averagely aggregating multiple images (inter-image) results in the determinative images being overwhelmed by other non-significant images.

\end{enumerate}

To alleviate the above problems, we propose a novel Information-Coupled Prototype Elaboration (ICPE) network for FSOD. As Figure~\ref{Figure 1}(b) shows, it generates representative and specific prototypes for each query image. Specifically, when encountering novel objects, humans intuitively apply the immediate information to search for the relevant knowledge they have seen for identifying the objects. Thus, a conditional information coupling module is proposed. It couples the information of query images into support images to generate coupled features with query-aware information, which are used to obtain tailored prototypes for each query image. In addition, the averaging operation might blur discriminative details that are crucial for object detection \cite{gao2019lip}. To address the issue, we propose a prototype dynamic aggregation module. It dynamically adjusts intra- and inter-image aggregation weights to highlight salient and useful information for predicting query images. On the one hand, an intra-image dynamic aggregation mechanism is designed to highlight the important local information within each support image by building local-to-global dependencies. On the other hand, based on the law of large numbers, an inter-image dynamic aggregation mechanism is proposed to learn the implicit similarities between support images and query images, with emphasis on highlighting the key support images. Through the above two modules, our model generates representative and tailored prototypes for each query image and effectively improves the performance.

The major contributions of this work are as follows: (1) A conditional information coupling module is proposed to couple query features into support features. It enhances support features with query-aware information. (2) We design a prototype dynamic aggregation module to emphasize the salient information both within each support image and cross multiple images when aggregating the enhanced support features into prototypes. (3) Extensive experiments illustrate that our method outperforms state-of-the-art methods in most experimental settings.

\begin{figure*}[t]
	\centering
	\includegraphics[width=0.9\textwidth]{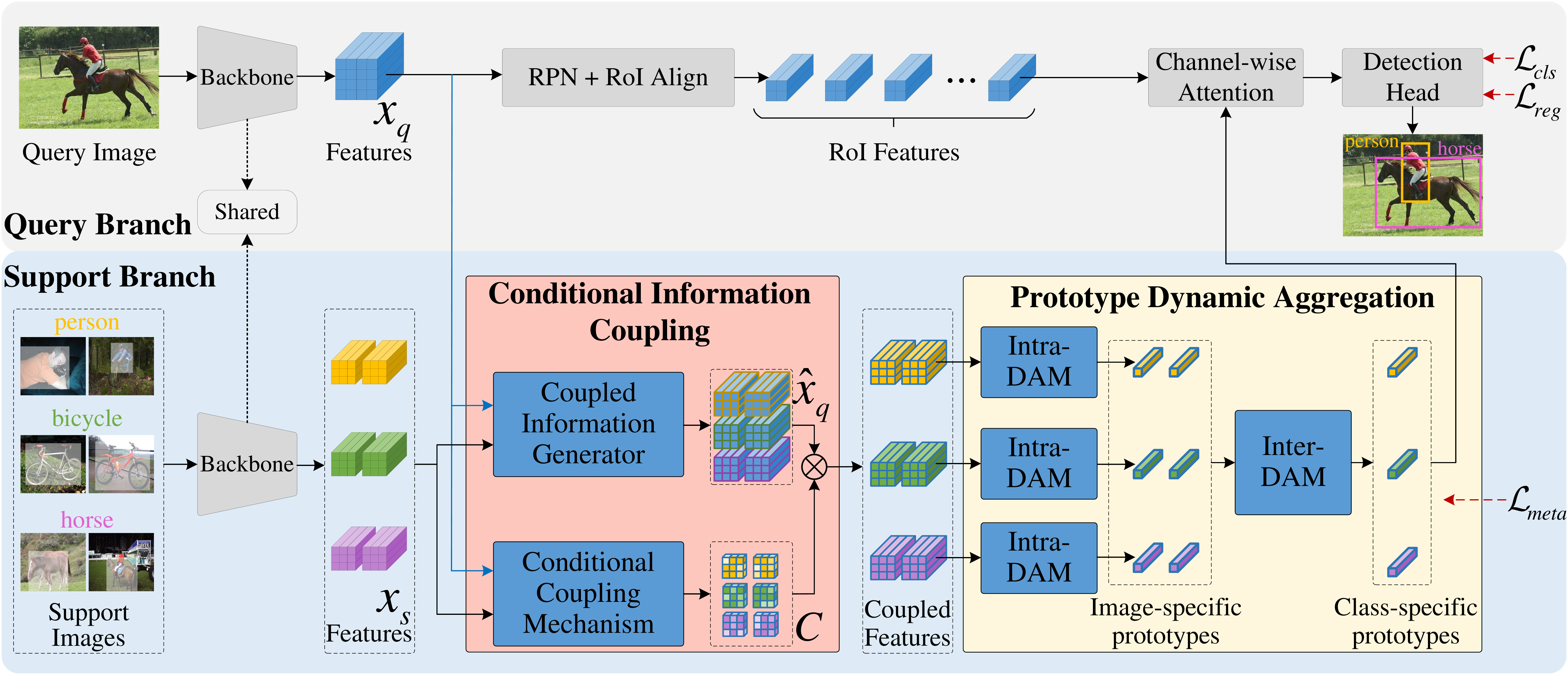}
	\caption{Architecture of ICPE. The conditional information coupling module integrates support features and query features to generate coupled features containing query-aware information. Through the prototype dynamic aggregation module, the coupled features are aggregated into representative prototypes which facilitate the query branch to detect objects accurately.}
	\label{Figure 2}
\end{figure*}

\section{Related Work}
\subsubsection{Object Detection.} Modern detectors are divided into one-stage and two-stage methods. The former \cite{redmon2016you,liu2016ssd,lin2017focal} directly predicts objects with a simple convolutional neural network, while the latter \cite{girshick2014rich,girshick2015fast,ren2015faster,he2017mask} extracts proposals on different receptive fields \cite{lin2017feature,mao2022beyond} and then makes refined predictions. Two-stage methods are generally more accurate than one-stage methods with slightly longer time. These detectors all rely on sufficient labeled data and suffer from over-fitting when facing limited data.

\subsubsection{Few-Shot Learning (FSL).} FSL is divided into augmentation, metric learning, and meta-learning methods. Augmentation based methods \cite{hariharan2017low,wang2018low} perform data or feature augmentation to meet the data requirements for training. Metric learning based methods \cite{koch2015siamese,scott2018adapted,mao2022bidirectional} calculate the distance between test and known samples to find adjacent classes. Meta-learning based methods \cite{andrychowicz2016learning,finn2017model,jamal2019task} construct few-shot tasks and learn meta-knowledge to guide models to adapt for new tasks quickly. These methods only focus on classification without locating objects.

\subsubsection{Few-Shot Object Detection (FSOD).} FSOD consists of data augmentation based, transfer learning based, and meta-learning based methods. The methods based on data augmentation \cite{yang2020context,wu2020multi,wu2021universal,xu2021few,zhang2021hallucination} alleviate data scarcity by expanding data or extracting various features from limited data. The transfer learning based methods \cite{chen2018lstd,karlinsky2019repmet,fan2020few,fan2021generalized} utilize fine-tuning or metric learning to transfer the knowledge learned on a data-rich base dataset to a few-shot novel dataset. TFA \cite{wang2020frustratingly} only fine-tunes the last layers of prediction heads on novel classes while freezing all other layers. The meta-learning based methods \cite{xiao2020few,hu2021dense,li2021beyond,li2021transformation,han2022meta} intuitively and explicitly simulate few-shot scenarios including query sets and support sets, and apply a query branch and a support branch to process the two sets, respectively. Based on YOLOv2 \cite{redmon2017yolo9000}, Meta YOLO \cite{kang2019few} extracts reweighting vectors from support images and applies them as channel-wise attention on query images to detect objects. Meta R-CNN \cite{yan2019meta} extracts prototypes and uses them to the region of interest (RoI) features of query images for detecting objects, which is selected as our baseline.

Although existing meta-learning methods have achieved prominent performance, the prototypes generated by them are immutable and lack query-aware information. We propose a method to generate unique prototypes for each query image and improve the quality of prototypes.

\section{Preliminaries}

\subsubsection{Problem Definition.} Given a base class set $ C_{b} $ with a data-rich dataset $ \mathcal D_{b} $ and a novel class set $ C_{n} $ ($ C_{n}\cap C_{b} = \oslash $) with a dataset $ \mathcal D_{n} $ which comprises $ k $ instances of each category, FSOD expects that detectors can detect objects of both $ C_{b} $ and $ C_{n} $ simultaneously. As with most progresses of FSOD \cite{li2021beyond,wu2020multi,xiao2020few}, we follow the episodic meta-learning paradigm \cite{kang2019few,yan2019meta}. It constructs abundant few-shot tasks with support sets and query sets in $ \mathcal D_{b} $ and $ \mathcal D_{n} $, respectively, namely episodes. And a two-step training strategy is applied. Firstly, the detector is trained on the episodes of $ \mathcal D_{b} $, called meta-training step. Then, in meta-finetuning step, the detector is fine-tuned on the episodes of $ \mathcal D_{n} $ and a subset of $ \mathcal D_{b} $.

\subsubsection{Baseline Method.} Meta R-CNN \cite{yan2019meta}, as a typical meta-learning based FSOD method, is selected as our baseline. Based on Faster R-CNN \cite{ren2015faster}, it contains a support branch and a query branch. In the support branch, images and object masks of support sets are sent to the backbone to obtain features, and global average pooling is applied to get feature vectors. Prototypes are obtained by averaging vectors belonging to the same class. The query branch receives query images and extracts RoI features. Then, given the generated prototypes, the channel-wise attention is applied for all RoIs to strengthen the category-related representations, promoting the detection head to detect objects. The prototypes extracted in Meta R-CNN are immutable and unrepresentative, while our model generates representative and tailored prototypes for query images.

\section{Method}
In this section, we first briefly introduce the architecture and the training strategy of our method. Then, we separately elaborate on the two proposed modules.

\subsection{Overview}

The purpose of our model is definite: Coupling query-perceptual information into support features and emphasizing salient details to generate adaptive high-quality prototypes. As Figure~\ref{Figure 2} shows, based on Meta R-CNN \cite{yan2019meta}, our model has a query branch and a support branch that share a common backbone, and introduces a conditional information coupling module and a prototype dynamic aggregation module for refining prototypes to be representative and tailored for each query image.

Given a query image and $ k $ support instances of each category ($ k $-shot), the features extracted from the backbone of the query branch and the support branch are sent to the conditional information coupling module. It effectively interacts the information of the two features to polish up support features with query-perceptual information. Then, the prototype dynamic aggregation module is introduced to highlight the salient information when aggregating the coupled features into class-specific prototypes. By introducing perceptual information of the query image and highlighting salient information, our model generates high-quality and representative prototypes tailored to the query image. Finally, the refined prototypes are used as the channel-wise attention for RoI features in the query branch, stimulating the detection head to accurately predict objects in the query image.

Like Meta R-CNN \cite{yan2019meta}, in addition to the typical detection loss consisting of classification loss $ \mathcal L_{cls} $ implemented by Cross-Entropy (CE) loss and regression loss $ \mathcal L_{reg} $ implemented by L1 loss, a meta loss $ \mathcal L_{meta} $ implemented by CE loss is also applied on the class-specific prototypes to make prototypes distinguishable and avoid prediction ambiguity. Thus, the overall loss of our ICPE is:
\begin{equation}
	\label{Eq 1}
	\mathcal L=\mathcal L_{cls}+\mathcal L_{reg}+\lambda \mathcal L_{meta}
\end{equation}
where $ \lambda $ is a loss weight for the balance between detection loss and meta loss, which is set to 1 in our experiments.

\begin{figure}[t]
	\centering
	\includegraphics[width=\columnwidth]{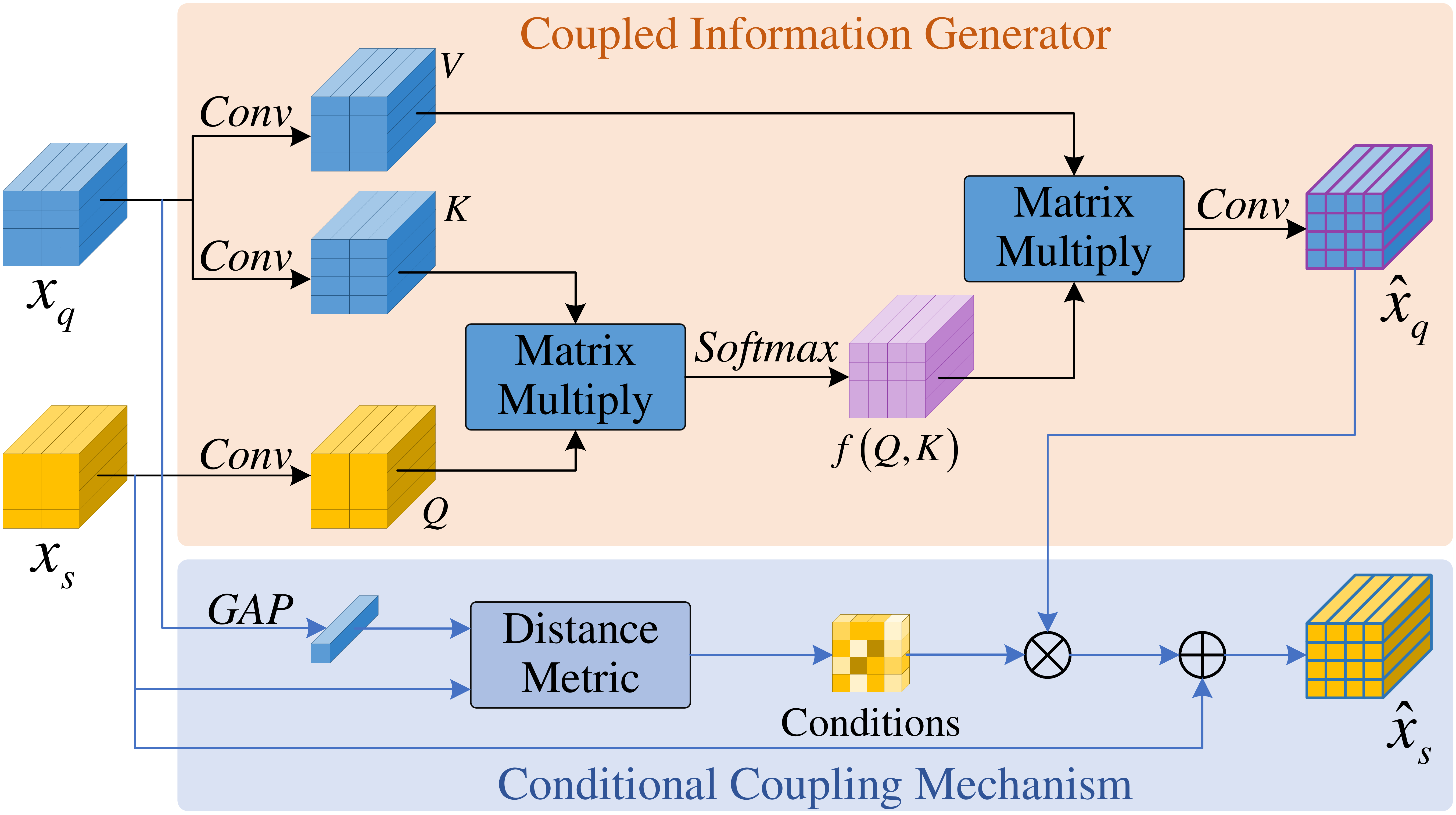}
	\caption{Architecture of the conditional information coupling module.}
	\label{Figure 3}
\end{figure}

\subsection{Conditional Information Coupling}

To make prototypes have query-perceptual information and be consistent with the objects in query images, QA-FewDet \cite{han2021query} applies heterogeneous graph networks to obtain query-adaptive prototypes. CoAE \cite{hsieh2019one} enriches query-aware information in support features through non-local \cite{wang2018non}. In order to avoid redundant and useless information flow when directly using non-local interaction, it is necessary to calculate the coupling conditions to guide the way of information coupling. Thus, as Figure~\ref{Figure 3} shows, we design a conditional information coupling module. It consists of a coupled information generator and a conditional coupling mechanism.

\textbf{Coupled information generator} polishes query features to be appropriate for coupling to support features. The query features are reorganized to obtain customized perceptual information for support features. For comprehensive information interaction, the coupled information generator is designed based on non-local, which measures similarities between the two features and realizes the recombination of query features. It receives features extracted from the query branch and the support branch, denoted as $ x_{q} $ and $ x_{s} $, respectively. $ x_{q} $ is represented as key-value pairs, and $ x_{s} $ is represented as query. The output is the restructured query features implemented as a weighted sum of values $ V $, where the weights are represented by the affinity relationship between $ x_{q} $ and $ x_{s} $, which is calculated by a compatibility function \cite{vaswani2017attention} of query $ Q $ and key $ K $:
\begin{equation}
	\label{Formulas 2}
	f\left ( Q,K \right )=Softmax\left ( Q^{T}\cdot K \right )
\end{equation}

Then, the values $ V $ of query features are assembled by the weights $ f\left ( Q,K \right ) $ and passed through a convolutional layer to produce the customized query features $ \hat{x}_{q} $ for coupling into support features:
\begin{equation}
	\hat{x}_{q}=Conv\left ( f\left ( Q,K \right )\cdot V \right )
\end{equation}

\textbf{Conditional coupling mechanism} calculates the conditions for information coupling based on $ x_{q} $ and $ x_{s} $. It predicts the regions of support features that are highly consistent with query features as coupling conditions, which guide the restructured query features $ \hat{x}_{q} $ to flow into support features $ x_{s} $ efficiently. Taking the global representation of $ x_{q} $, the conditional coupling mechanism adequately delves the areas of $ x_{s} $ with high similarities of $ x_{q} $. The conditions $ C $ are calculated as follows:
\begin{equation}
	C=\mathcal M \left ( GAP\left ( x_{q} \right ),x_{s} \right )
\end{equation}
where $ GAP $ denotes the global average pooling, and $ \mathcal M $ represents the distance measurement implemented with cosine similarity. Then, $ C $ is used as a mask to activate the areas of $ \hat{x}_{q} $ for information coupling. And the information coupling from query features to support features is realized by a residual connection \cite{he2016deep}. Thus, the coupled features $ \hat{x}_{s} $ with query-perceptual information are expressed as follows:
\begin{equation}
	\hat{x}_{s}=C\circ \hat{x}_{q}+x_{s}
\end{equation}
where $ \circ $ represents Hadamard product. Compared with the original support features, $ \hat{x}_{s} $ contains the perceptual information of query features through the deep and efficient interaction between the two features.

Different from DCNet \cite{hu2021dense} which replaces channel-wise attention in the query branch with cross-attention to densely match support features (prototypes) into query features, our coupling module performs cross-attention to improve the quality of prototypes. It enhances query-aware information in support features which are used to obtain unique and tailored prototypes for each query image subsequently. Furthermore, our module extends cross-attention with a conditional coupling mechanism for efficient information coupling and reduced noise introduction.

\begin{figure}[t]
	\centering
	\includegraphics[width=\columnwidth]{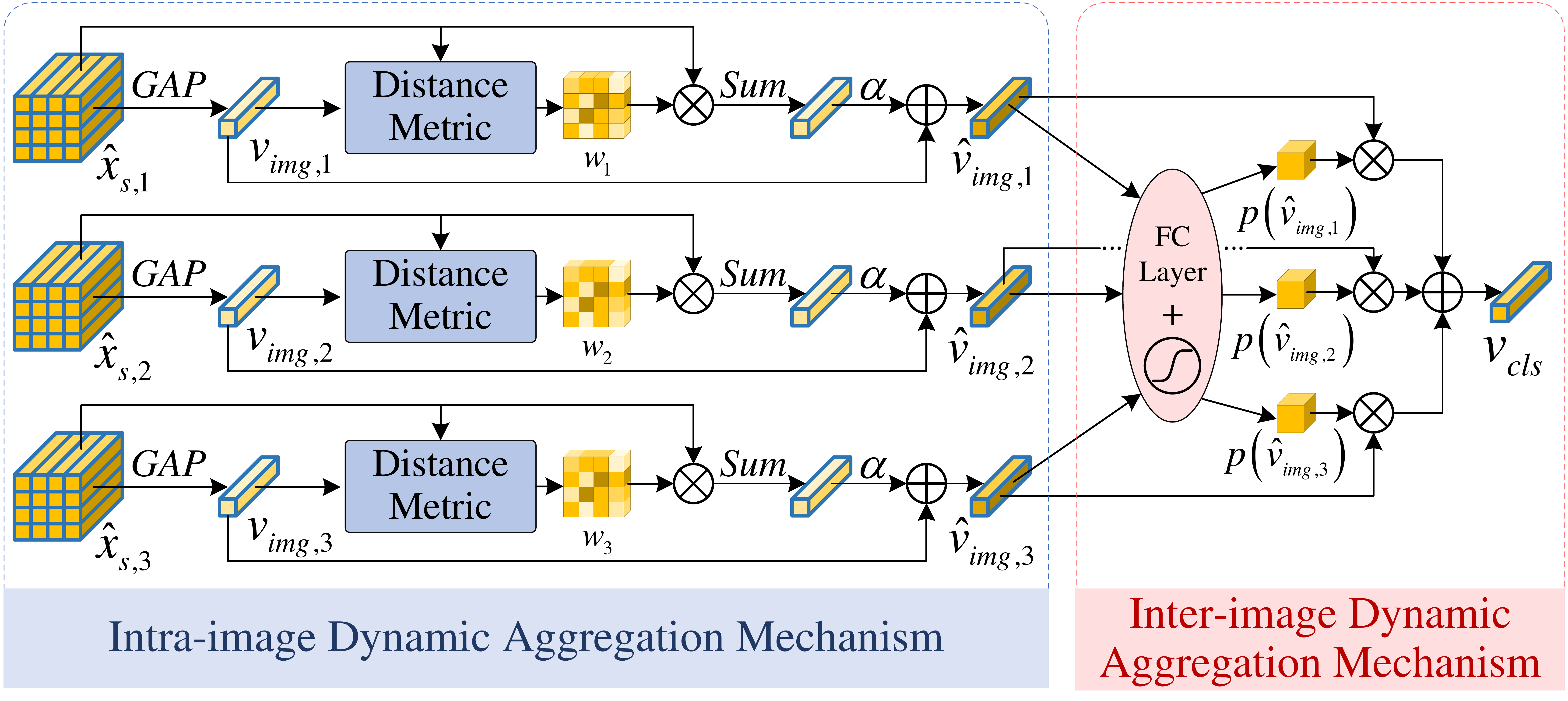}
	\caption{Architecture of the prototype dynamic aggregation module.}
	\label{Figure 4}
\end{figure}

\subsection{Prototype Dynamic Aggregation}
For dynamic aggregation of support features, CrossTransformers \cite{doersch2020crosstransformers} aggregates information in a spatially-aware way. DAnA \cite{chen2021dual} views features as a wave along the channel dimension and maintains the wave patterns of foreground features. To generate high-quality and representative class-specific prototypes from $ \hat{x}_{s} $, a prototype dynamic aggregation module is proposed. As Figure~\ref{Figure 4} shows, it incorporates intra-image and inter-image dynamic aggregation mechanisms. They replace average aggregations both within a single image and across multiple images, preserving the discriminative and salient details when aggregating features into prototype vectors.

\textbf{Intra-image dynamic aggregation mechanism (Intra-DAM)} is designed to preserve prominent features within each coupled feature by building local-to-global dependencies when generating image-specific prototypes. Unlike generating prototypes with GAP, Intra-DAM reinforces the contributions of salient regions. It conducts the global representation of $ \hat{x}_{s} $, and then calculates the global dependency of each local pixel to obtain the dynamic aggregation weights within each coupled feature. The weight $ w $ signifies the amount of the global information representing the whole features contained in each pixel, which is defined as:
\begin{equation}
	w=\mathcal N \left ( GAP\left ( \hat{x}_{s} \right ),\hat{x}_{s} \right )
\end{equation}
where $ \mathcal N $ means a function that conducts the dependency between two features, which is implemented with the cosine similarity. After yielding the aggregation weight of each position, the pixels of $ \hat{x}_{s} $ are aggregated to generate a refined and representative image-specific prototype $ \hat{v}_{img} $:
\begin{equation}
	\hat{v}_{img}=v_{img} + \frac{\alpha }{N} \sum w\circ \hat{x}_{s}
\end{equation}
where $ v_{img}=GAP\left ( \hat{x}_{s} \right )$ is a popular way to generate prototypes. $ \sum $ denotes the sum operation of vectors. $ N $ means the number of vectors for normalization, and $ \alpha $ is a balance factor set to 1. The generated $ \hat{v}_{img} $ emphasizes salient pixels within $ \hat{x}_{s} $ on the basis of treating all pixels equally.

\textbf{Inter-image dynamic aggregation mechanism (Inter-DAM)} achieves dynamic aggregation between multiple image-specific prototypes of the same class to generate class-specific prototypes. For a query image, the discriminative information carried in different support images is diverse. Inter-DAM consolidates the dominance of the coupled features which contain relatively rich salient and useful information for detection. Given the image-specific prototypes of each class, it models the implicit similarities between multiple support images and the query image through a fully connected (FC) layer to generate contribution probabilities $ p\left ( \hat{v}_{img} \right ) $, that is, the proportion of the useful information carried in each support image, which is defined as:
\begin{equation}
	p\left ( \hat{v}_{img} \right ) = Sigmoid\left ( FC\left ( \hat{v}_{img} \right ) \right )
\end{equation}

Based on the contribution probabilities $ p\left ( \hat{v}_{img} \right ) $ of support images, the weighted summation is applied to $ \hat{v}_{img} $ belonging to the same category to get the expectation of them, which is denoted as the class-specific prototype:
\begin{equation}
	v_{cls}=E\left ( \hat{v}_{img} \right )=\sum_{i=1}^{k}p\left ( \hat{v}_{img,i} \right )\hat{v}_{img,i}
\end{equation}
where $ k $ means the number of support instances for each class, i.e., $ k $-shot. As noted by the Law of Large Numbers, the barycenter of samples converges to the expectation as the number of samples becomes infinity. Hence, with few-shot images, the class-specific prototypes represented by the expectations of image-specific prototypes can simulate the class information in the presence of sufficient data. The generated class-specific prototypes $ v_{cls} $ adequately carry the representative and salient information of categories.

\begin{table*}[t]
	\setlength{\tabcolsep}{3pt}
	\renewcommand{\arraystretch}{1.1}
	\begin{center}
		\begin{tabular}{c|ccccc|ccccc|ccccc}
			\toprule
			\multirow{2}{*}{Method / Shot} & \multicolumn{5}{c|}{Novel Set 1}   & \multicolumn{5}{c|}{Novel Set 2}    & \multicolumn{5}{c}{Novel Set 3}       \\ 
			& 1             & 2             & 3             & 5             & 10            & 1             & 2             & 3             & 5             & 10            & 1             & 2             & 3             & 5             & 10           \\ 
			\hline
			\hline
			LSTD \cite{chen2018lstd}                & 8.2           & 1.0           & 12.4          & 29.1          & 38.5          & 11.4          & 3.8           & 5.0           & 15.7          & 31.0          & 12.6          & 8.5           & 15.0          & 27.3          & 36.3            \\
			Meta YOLO \cite{kang2019few}           & 14.8          & 15.5          & 26.7          & 33.9          & 47.2          & 15.7          & 15.3          & 22.7          & 30.1          & 40.5          & 21.3          & 25.6          & 28.4          & 42.8          & 45.9            \\
			Meta R-CNN \cite{yan2019meta}          & 19.9          & 25.5          & 35.0          & 45.7          & 51.5          & 10.4          & 19.4          & 29.6          & 34.8          & 45.4          & 14.3          & 18.2          & 27.5          & 41.2          & 48.1               \\
			Viewpoint \cite{xiao2020few}           & 24.2          & 35.3          & 42.2          & 49.1          & 57.4          & 21.6          & 24.6          & 31.9          & 37.0          & 45.7          & 21.2          & 30.0          & 37.2          & 43.8          & 49.6           \\
			TFA w/fc \cite{wang2020frustratingly}           & 36.8          & 29.1          & 43.6          & 55.7          & 57.0          & 18.2          & 29.0          & 33.4          & 35.5          & 39.0          & 27.7          & 33.6          & 42.5          & 48.7          & 50.2              \\
			TFA w/cos \cite{wang2020frustratingly}           & 39.8          & 36.1          & 44.7          & 55.7          & 56.0          & 23.5          & 26.9          & 34.1          & 35.1          & 39.1          & 30.8          & 34.8          & 42.8          & 49.5          & 49.8              \\
			MPSR \cite{wu2020multi}         & 41.7          & 42.5          & 51.4          & 52.2          & 61.8 & 24.4          & 29.3          & 39.2          & 39.9          & 47.8 & 35.6          & 41.8          & 42.3          & 48.0          & 49.7            \\
			SRR-FSD \cite{zhu2021semantic}     & 47.8          & 50.5          & 51.3          & 55.2          & 56.8          & 32.5          & 35.3          & 39.1          & 40.8          & 43.8          & 40.1          & 41.5          & 44.3          & 46.9          & 46.4              \\
			FSCE \cite{sun2021fsce}  & 44.2          & 43.8          & 51.4          & 61.9          & 63.4          & 27.3          & 29.5          & 43.5          & 44.5          & 50.2          & 37.2          & 41.9          & 47.5          & 54.6          & 58.5            \\
			DCNet \cite{hu2021dense}               & 33.9          & 37.4          & 43.7          & 51.1          & 59.6          & 23.2          & 24.8          & 30.6          & 36.7          & 46.6          & 32.3          & 34.9          & 39.7          & 42.6          & 50.7                \\
			CME \cite{li2021beyond}                 & 41.5          & 47.5          & 50.4          & 58.2 & 60.9          & 27.2          & 30.2          & 41.4          & 42.5          & 46.8          & 34.3          & 39.6          & 45.1          & 48.3          & 51.5        \\
			DeFRCN \cite{qiao2021defrcn}  & \underline{53.6}          & \underline{57.5}          & \underline{61.5}          & 64.1          & 60.8          & 30.1          & \underline{38.1}          & \underline{47.0}  & \textbf{53.3}          & 47.9          & \underline{48.4}          & \underline{50.9}          & 52.3          & 54.9          & 57.4            \\
			Meta FRCNN \cite{han2022meta}    & 43.0          & 54.5          & 60.6          & \textbf{66.1} & 65.4          & 27.7          & 35.5          & 46.1          & 47.8          & \underline{51.4}         & 40.6          & 46.4          & \underline{53.4}          & \underline{59.9}          & \underline{58.6}  \\
			KFSOD \cite{zhang2022kernelized}   & 44.6     & -     & 54.4          & 60.9          & \underline{65.8}          & \textbf{37.8}          & -          & 43.1          & 48.1          & 50.4          & 34.8          & -          & 44.1          & 52.7          & 53.9      \\
			FCT \cite{han2022few}  & 38.5     & 49.6     & 53.5          & 59.8          & 64.3          & 25.9          & 34.2          & 40.1          & 44.9          & 47.4          & 34.7          & 43.9          & 49.3          & 53.1          & 56.3      \\
			\hline
			\textbf{ICPE(ours)} & \textbf{54.3} & \textbf{59.5} & \textbf{62.4} & \underline{65.7}          & \textbf{66.2}    & \underline{33.5} & \textbf{40.1} & \textbf{48.7} & \underline{51.7} & \textbf{52.5}          & \textbf{50.9} & \textbf{53.1} & \textbf{55.3} & \textbf{60.6} & \textbf{60.1}      \\ 
			\bottomrule
		\end{tabular}
		\caption{FSOD performance in terms of mAP on Pascal VOC dataset. Best results are in \textbf{bold} and second best are \underline{underlined}.}
		\label{Table 1}
	\end{center}
\end{table*}

\begin{table}[t]
	\setlength{\tabcolsep}{1.5pt}
	\renewcommand{\arraystretch}{1.1}
	\begin{center}
		\begin{tabular}{c|ccc|ccc}
			\toprule
			\multirow{2}{*}{Method / Shot} & \multicolumn{3}{c|}{10}   & \multicolumn{3}{c}{30}   \\ 
			& $ AP $   & $ AP_{50} $ & $ AP_{75} $ & $ AP $   & $ AP_{50} $ & $ AP_{75} $  \\ 
			\hline
			\hline
			LSTD \shortcite{chen2018lstd}      & 3.2  & 8.1  & 2.1 & 6.7  & 15.8 & 5.1    \\
			Meta YOLO \shortcite{kang2019few}  & 5.6  & 12.3 & 4.6 & 9.1  & 19.0 & 7.6    \\
			Meta R-CNN \shortcite{yan2019meta} & 8.7  & 19.1 & 6.6& 12.4 & 25.3 & 10.8  \\
			Viewpoint \shortcite{xiao2020few} & 12.5 & \underline{27.3} & 9.8  & 14.7 & 30.6 & 12.2   \\
			TFA w/fc \shortcite{wang2020frustratingly}  & 10.0  & 19.2 & 9.2   & 13.4  & 24.7 & 13.2   \\
			TFA w/cos \shortcite{wang2020frustratingly}  & 10.0  & 19.1 & 9.3   & 13.7 & 24.9 & 13.4    \\
			MPSR \shortcite{wu2020multi}       & 9.8  & 17.9 & 9.7   & 14.1 & 25.4 & 14.2 \\
			SRR-FSD \shortcite{zhu2021semantic}  & 11.3  & 23.0 & 9.8  & 14.7  & 29.2 &  13.5   \\
			FSCE \shortcite{sun2021fsce}  & 11.1  & - & 9.8 & 15.3  & - &  14.2      \\
			DCNet \shortcite{hu2021dense}      & 12.8 & 23.4 & 11.2  & 18.6 & \underline{32.6} & 17.5  \\
			CME \shortcite{li2021beyond}        & 15.1 & 24.6 & 16.4  & 16.9 & 28.0 & \underline{17.8}  \\
			DeFRCN \shortcite{qiao2021defrcn}   & \underline{18.5}  & - & -  & \underline{22.6}  &  - &  -   \\
			Meta FRCNN \shortcite{han2022meta}     & 12.7  & 25.7 & 10.8    & 16.6  & 31.8  &  15.8  \\
			KFSOD \shortcite{zhang2022kernelized}    & \underline{18.5}  & 26.3 &  \textbf{18.7}  & -  & - &   -   \\
			FCT \shortcite{han2022few}    & 15.3  & - &  -  & 20.2  & - & -    \\
			
			\hline
			\textbf{ICPE(ours)} & \textbf{19.3} & \textbf{27.9} & \underline{18.0}  & \textbf{23.1} & \textbf{32.9} & \textbf{19.2}   \\ 
			\bottomrule
		\end{tabular}
		\caption{FSOD performance on MS COCO dataset.}
		\label{Table 2}
	\end{center}
\end{table}

\begin{table}[t]
	\setlength{\tabcolsep}{3pt}
	\renewcommand{\arraystretch}{1.1}
	\begin{center}
		\begin{tabular}{c|cc|ccccc|cc}
			\toprule
			& \multicolumn{2}{c|}{Method} &   \multicolumn{5}{c|}{Shot} &   &    \\
			\hline
			\# & CIC          & PDA          & 1    & 2    & 3    & 5    & 10   & $ Avg. $  & $ \Delta Avg. $\\ 
			\hline
			\hline
			1  &              &                  & 44.7 & 49.3 & 55.5 & 58.3 & 61.2   & 53.8  &   \\
			2  &  $ \checkmark $  &              & 50.7 & 57.2 & 59.6 & 63.2 & 64.1 &  59.0 & +5.2 \\
			3  &              &  $ \checkmark $  & 49.3 & 55.9 & 57.3 & 61.4 & 63.8 & 57.5  & +3.7   \\
			4  &  $ \checkmark $  &  $ \checkmark $  & \textbf{54.3} & \textbf{59.5} & \textbf{62.4} & \textbf{65.7} & \textbf{66.2} &  \textbf{61.6} & \textbf{+7.8}  \\
			\bottomrule
		\end{tabular}
		\caption{Ablation study of different modules. CIC means the conditional information coupling module. PDA means the prototype dynamic aggregation module. $ Avg. $ represents the average performance of all shot settings.}
		\label{Table 3}
	\end{center}
\end{table}

\section{Experiments}
\subsection{Experimental Setup}
\subsubsection{Dataset.} Same as Meta YOLO \cite{kang2019few} and Meta R-CNN \cite{yan2019meta}, our method is evaluated on Pascal VOC \cite{everingham2010pascal,everingham2015pascal} and MS COCO \cite{lin2014microsoft} datasets. For Pascal VOC, our model is trained on the trainval sets of VOC 2007 and 2012, and tested on VOC 2007 test set. The dataset is partitioned into three different splits, where 5 categories are selected as novel classes and the other 15 categories are used as base classes. There are $ k $ annotated instances of each category for meta-finetuning. $ k $ is set to 1, 2, 3, 5, and 10. For MS COCO, the model is trained on a modified dataset consisting of 80k training images and 35k validation images. The remaining 5k images are used for test. The 20 classes included in Pascal VOC are used as novel classes, and the other 60 categories are as base classes. $ k $ is set to 10 and 30.

\subsubsection{Implementation Details.} Same as previous work \cite{wang2020frustratingly,qiao2021defrcn}, for query images, we use multiple scale images for training and a single scale for testing. The shorter sides are $ 480\sim 800 $ pixels while the longer sides are less than $ 1333 $ pixels. The support images are resized to $ 224\times 224 $ pixels. Only random horizontal flipping and normalization are used for training. The backbone is ResNet-101 \cite{he2016deep} with the pretrained weight on ImageNet \cite{russakovsky2015imagenet}. We set batch size as 4 and use stochastic gradient descent (SGD) optimizer with the momentum of $ 0.9 $ and the weight decay of $ 0.0001 $. The training iterations and learning rates are the same as Meta R-CNN \cite{yan2019meta}. Consistent with TFA \cite{wang2020frustratingly}, the present results of our method are averaged over multiple random runs. We implement our experiments in PyTorch \cite{paszke2019pytorch} on 2 V100 GPUs.

\subsection{Comparisons with State-of-the-art Methods}
We conduct experiments on Pascal VOC and MS COCO, and compare our method with state-of-the-art (SOTA) methods. The results demonstrate the effectiveness of our ICPE.

\subsubsection{Pascal VOC.} In Table~\ref{Table 1}, we report the results of our method and existing SOTA methods on Pascal VOC. Our ICPE achieves state-of-the-art performance on almost all three splits with different shot settings. For the average mAP of three splits, our method achieves 46.2\%, 50.9\%, 55.5\%, 59.3\%, and 59.6\% in 1, 2, 3, 5, and 10-shot. As the number of annotated instances increases, the performance of the model gradually improves. Especially in extremely low-shot settings, the performance is improved significantly as the number of shots increases. For example, when only increasing one instance from 1-shot to 2-shot and from 2-shot to 3-shot, the average performance of all splits is improved by 4.7\% and 4.6\%, respectively. This is because when the number of samples is tiny, adding one sample can provide relatively rich information for prototypes.

\begin{figure}[t]
	\centering
	\includegraphics[width=\columnwidth]{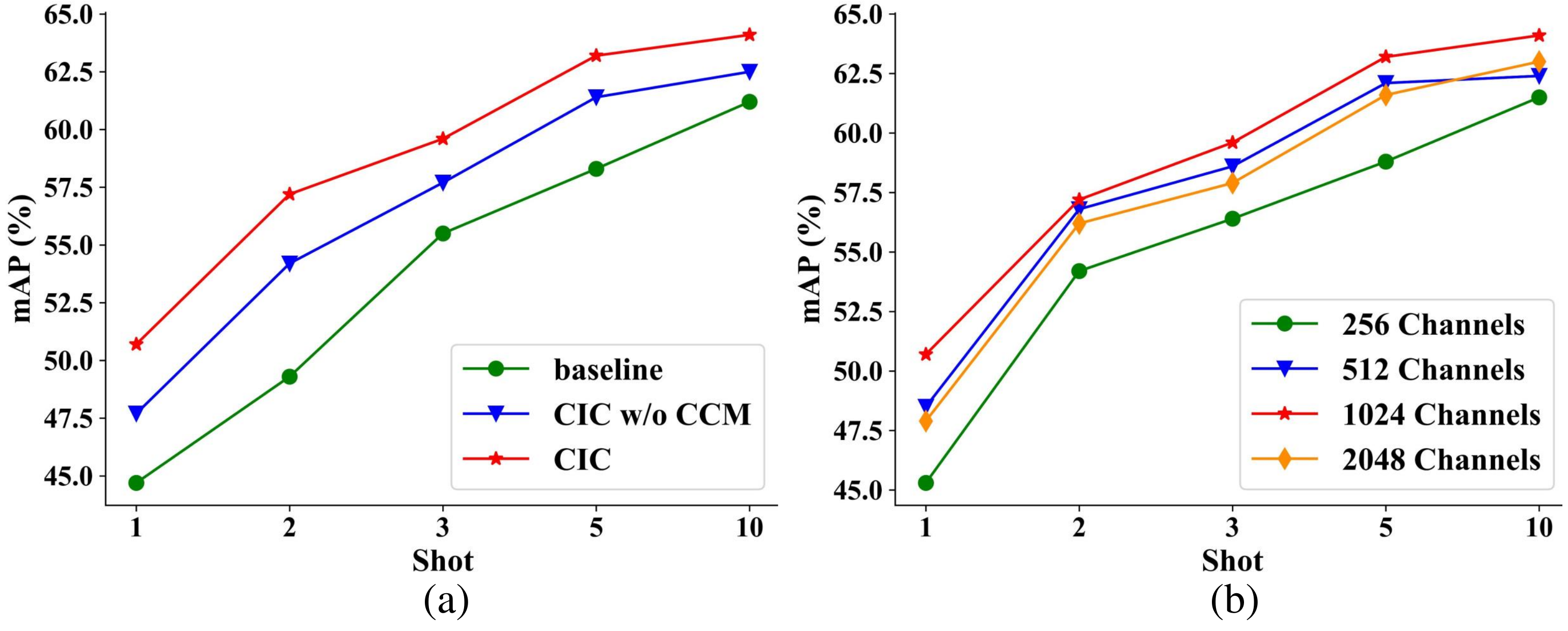}
	\caption{Ablation study of Conditional Information Coupling (CIC). (a) Effects of conditional coupling mechanism (CCM). (b) Effects of the number of channels in $ f\left ( Q,K \right ) $.}
	\label{Figure 5}
\end{figure}

\begin{table}[t]
	\setlength{\tabcolsep}{3pt}
	\renewcommand{\arraystretch}{1.1}
	\begin{center}
		\begin{tabular}{c|ccccc|cc}
			\toprule
			&            &      &   Shot   &     &      &      &      \\ 
			\hline
			Method                   & 1    & 2    & 3    & 5    & 10   & $ Avg. $ & $ \Delta Avg. $ \\ 
			\hline
			\hline
			baseline    & 44.7 & 49.3 & 55.5 & 58.3 & 61.2 &  53.8   &    \\
			\hline
			+GMP       & 45.9 & 52.2 & 55.8 & 58.7 & 61.9 & 54.9   &   +1.1 \\
			+LIP       & 47.5 & 52.8 & 56.0 & 59.2 & 61.9 &  55.5   &  +1.7 \\
			+Intra-DAM & 49.3 & 54.4 & 56.4 & 59.9 & 62.6 & 56.5   &  +2.7 \\
			\hline
			\multicolumn{1}{c|}{\begin{tabular}[c]{@{}c@{}}+Intra-DAM \\ \&Inter-DAM\end{tabular}} & \textbf{49.3} & \textbf{55.9} & \textbf{57.3} & \textbf{61.4} & \textbf{63.8} &  \textbf{57.5}  & \textbf{+3.7} \\
			\bottomrule
		\end{tabular}
		\caption{Ablation study of prototype dynamic aggregation. +GMP means that image-specific prototypes are implemented by the sum of GAP and global max pooling (GMP). +LIP means generating image-specific prototypes via Local Importance-based Pooling \cite{gao2019lip}.}
		\label{Table 4}
	\end{center}
\end{table}

\begin{table}[!h]
	\setlength{\tabcolsep}{2.5pt}
	\renewcommand{\arraystretch}{1.1}
	\begin{center}
		\begin{tabular}{c|ccc}
			\toprule
			Method    & Parameters(MB) & FPS(img/s) & FLOPs(GB) \\
			\hline
			\hline
			baseline   & 45.9           & 13.4       & 1752.7    \\
			ICPE(ours) & 50.1           & 9.6        & 2401.1    \\
			\bottomrule
		\end{tabular}
		\caption{Ablation study of computational cost.}
		\label{Table 5}
	\end{center}
\end{table}

\subsubsection{MS COCO.} We also evaluate our method on a challenging dataset MS COCO. The results with the standard COCO metrics are reported in Table~\ref{Table 2}. Compared with Meta R-CNN \shortcite{yan2019meta}, our method improves $ AP $ by more than 10\% in both 10 and 30-shot settings, reaching 19.3\% and 23.1\%, respectively. Besides, our method outperforms SOTA methods in most evaluation metrics. It shows that our ICPE is still effective on the challenging and complex datasets.

\subsection{Ablation Study}
To analyze the effectiveness of the two components proposed in ICPE, we conduct comprehensive ablation studies on the Novel Set 1 of Pascal VOC. And the baseline is reproduced to verify the superiority of our method convincingly.

\subsubsection{Effects of Different Modules.} Table~\ref{Table 3} shows the effectiveness of the proposed modules in ICPE. The conditional information coupling module improves the average performance of all shot settings by 5.2\%. It augments query-aware information in support features. Besides, the average performance gain is 3.7\% when simply introducing the prototype dynamic aggregation module to highlight useful and salient information. With both conditional information coupling and prototype dynamic aggregation, the performance gain increases to 7.8\%. Thus, our ICPE efficiently strengthens the query-aware and salient information when generating high-quality prototypes tailored to query images. It achieves significant improvement over the baseline.

\subsubsection{Effects of Conditional Information Coupling.} Figure~\ref{Figure 5}(a) verifies the advancement of the conditional coupling mechanism. When only using the coupled information generator (CIC w/o CCM), the mAPs of five settings are 47.7\%, 54.2\%, 57.7\%, 61.4\%, and 62.5\%, respectively. The average performance is 2.9\% higher than the baseline. Further adding the conditional coupling mechanism, as \#2 in Table~\ref{Table 3} shows, the mAP is improved by 5.2\%. Besides, Figure~\ref{Figure 5}(b) reflects the effects of the number of channels in $ f\left ( Q,K \right ) $. It reveals that $ f\left ( Q,K \right ) $ with 1024 channels reports the best results. 256 and 512 channels are insufficient because of the depression of feature representation, while 2048 channels face performance degradation due to over-fitting.

\subsubsection{Effects of Prototype Dynamic Aggregation.} To verify the effectiveness of Intra-DAM, we conduct experiments with two other approaches to generate image-specific prototypes: one is to use the sum of GAP and global max pooling (GMP), and the other is to use LIP \cite{gao2019lip}. As Table~\ref{Table 4} shows, the performance with GMP retaining the most discriminative features is increased by 1.1\%. And LIP, strengthening the local importance, improves the performance by 1.7\%. Our Intra-DAM improves the performance by 2.7\%. Further adding Inter-DAM, the average performance is improved by 3.7\%. It is concluded that the prototype dynamic aggregation module effectively highlights salient and useful information when generating prototypes.

\subsubsection{Computational Cost.} We analyze the computational cost between our ICPE and Meta R-CNN on 10-shot settings. As Table~\ref{Table 5} shows, compared with the baseline, our model only increases a slight number of parameters and computations.

\begin{figure}[t]
	\centering
	\includegraphics[width=\columnwidth]{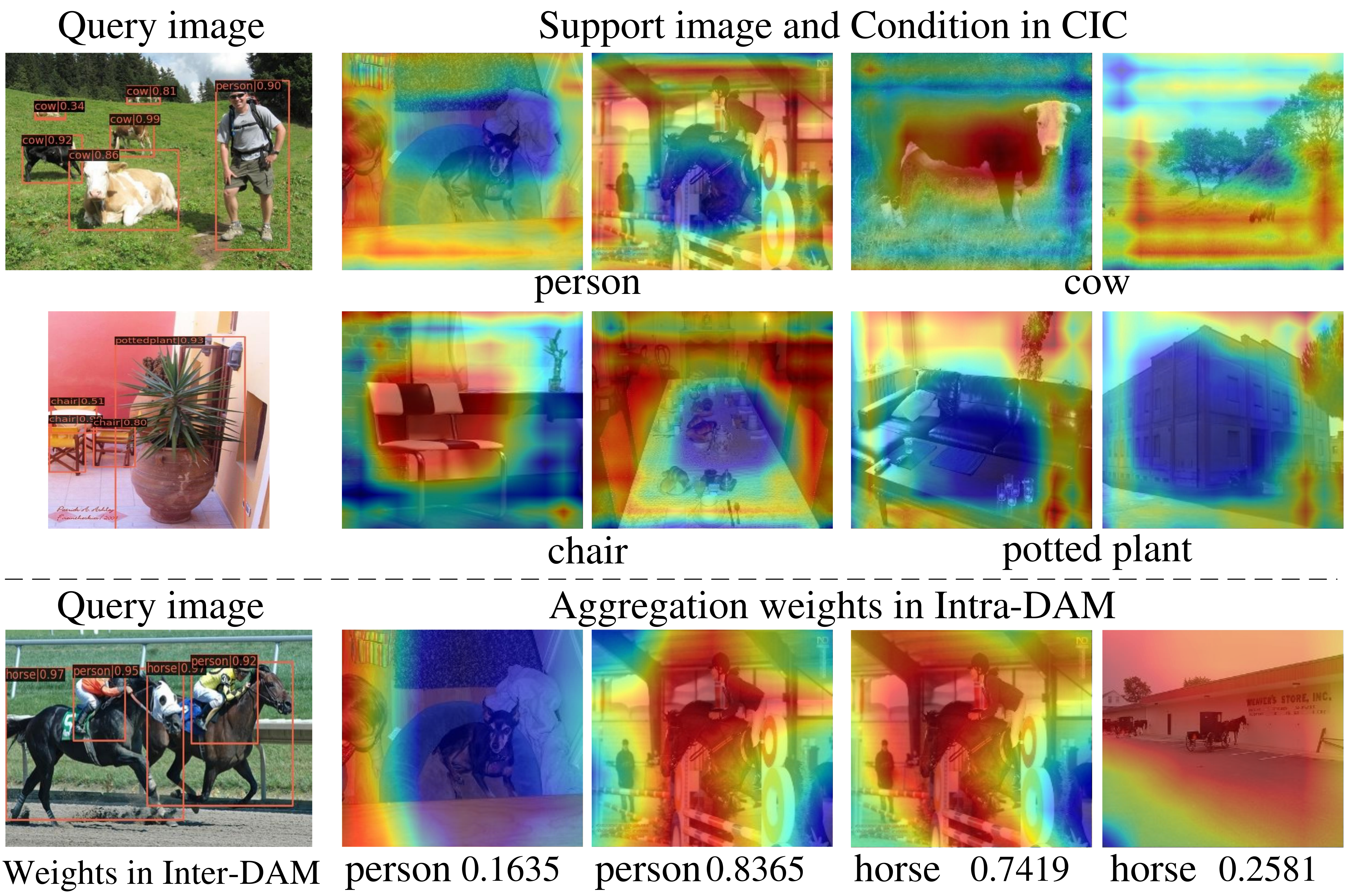}
	\caption{Visualization of components. Top: Conditions in conditional information coupling module. Bottom: Aggregation weights of prototype dynamic aggregation module.}
	\label{Figure 6}
\end{figure}

\begin{figure}[t]
	\centering
	\includegraphics[width=\columnwidth]{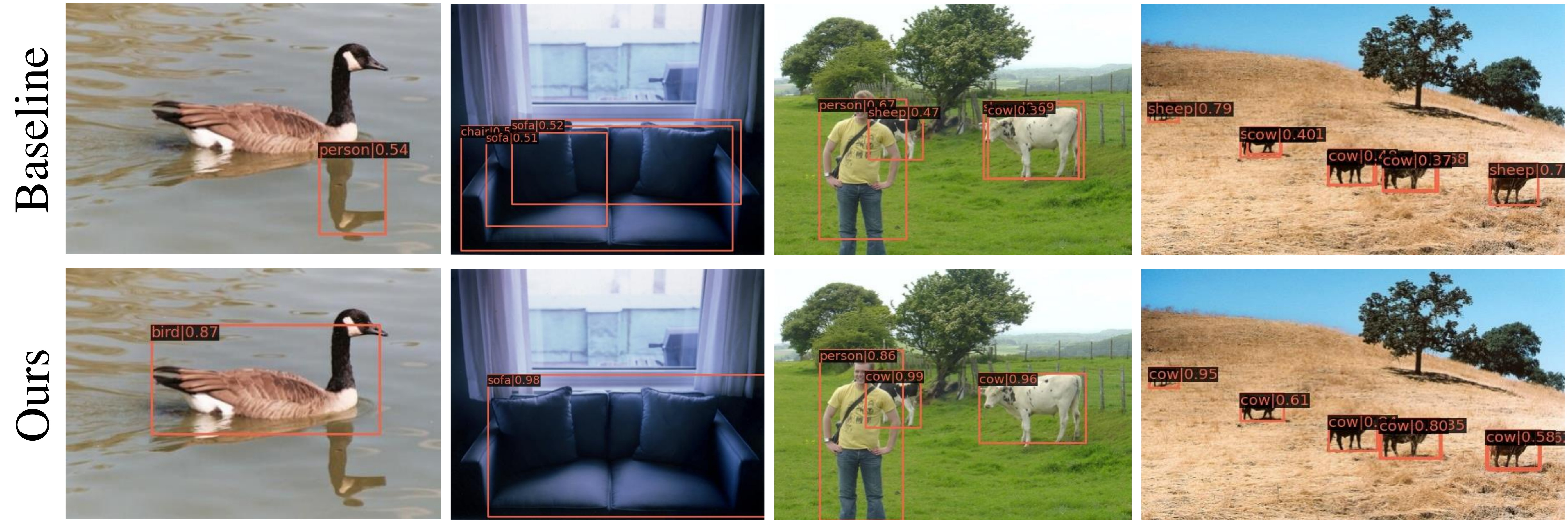}
	\caption{Detection results of the baseline and our method.}
	\label{Figure 7}
\end{figure}

\subsection{Qualitative Results}

In Figure~\ref{Figure 6}, the conditions predicted by the conditional coupling mechanism are almost the same as the areas where the objects are located in. It guides the perceptual information of query images to be coupled accurately to the effective regions of support images and reduces redundant information flow. Besides, for prototype dynamic aggregation module, Intra-DAM pays attention to the areas with comprehensive information in each support image, and Inter-DAM assigns significant weights to the images with high similarities for query images. We also present the detection results of our method and the baseline in Figure~\ref{Figure 7}. Our ICPE achieves superior performance and effectively reduces false detection.

\section{Conclusion}

In this work, we propose an Information-Coupled Prototype Elaboration network to generate high-quality prototypes tailored to each query image. A conditional information coupling module is proposed to enhance query-perceptual information in support features. Besides, we design a prototype dynamic aggregation mechanism to highlight the salient and useful information both within a single image and across multiple images when generating prototypes. Extensive experiments validate the effectiveness of our method which achieves state-of-the-art performance in most settings. 

\section{Acknowledgments}
This work was supported by the National Key R\&D Program of China (Grant No. 2021YFB390050*).

\bibliography{aaai23}

\end{document}